\documentclass[conference]{IEEEtran}
\IEEEoverridecommandlockouts
\usepackage{cite}
\usepackage{amsmath,amssymb,amsfonts}
\usepackage{algorithmic}
\usepackage{textcomp}
\usepackage{caption}

\usepackage{graphicx}
\usepackage{amsfonts}
\usepackage[ruled,vlined]{algorithm2e}
\usepackage{comment}
\usepackage{url}
\usepackage[dvipsnames]{xcolor}
\usepackage[caption=false]{subfig}
\usepackage{verbatim}
\newtheorem{definition}{Definition}
\def\anonymized{false}

\def\BibTeX{{\rm B\kern-.05em{\sc i\kern-.025em b}\kern-.08em
    T\kern-.1667em\lower.7ex\hbox{E}\kern-.125emX}}
\begin{document}

\title{Hidden Markov Model to Predict Tourists Visited Places}

\author{\IEEEauthorblockN{Theo Demessance}
\IEEEauthorblockA{\textit{Léonard de Vinci, Research Center, } \\
\textit{92 916 Paris La Défense, France}\\
\textit{College of Intelligence and Computing}\\
\textit{ Tianjin University}\\
theo.demessance@edu.devinci.fr}\\
\and
\IEEEauthorblockN{ Chongke Bi}
\IEEEauthorblockA{\textit{College of Intelligence and Computing}\\
\textit{ Tianjin University}\\
bichongke@tju.edu.cn}
\and
\IEEEauthorblockN{ Sonia Djebali}
\IEEEauthorblockA{\textit{Léonard de Vinci, Research Center, } \\
\textit{92 916 Paris La Défense, France}\\
sonia.djebali@devinci.fr}
\and
\IEEEauthorblockN{Guillaume Gu\'erard}
\IEEEauthorblockA{\textit{Léonard de Vinci, Research Center, } \\
\textit{92 916 Paris La Défense, France}\\
guillaume.guerard@devinci.fr}
}

\maketitle
\begin{abstract}
Nowadays, social networks are becoming a popular way of analyzing tourist behavior, thanks to the digital traces left by travelers during their stays on these networks.
The massive amount of data generated; by the propensity of tourists to share comments and photos during their trip; makes it possible to model their journeys and analyze their behavior.
Predicting the next movement of tourists plays a key role in tourism marketing to understand demand and improve decision support.

In this paper, we propose a method to understand and to learn tourists' movements based on social network data analysis to predict future movements. The method relies on a machine learning grammatical inference algorithm.
A major contribution in this paper is to adapt the grammatical inference algorithm to the context of big data.
Our method produces a hidden Markov model representing the movements of a group of tourists. The hidden Markov model is flexible and editable with new data.
The capital city of France, Paris is selected to demonstrate the efficiency of the proposed methodology.
\end{abstract}

\begin{IEEEkeywords}
Tourist behavior, Hidden Markov Model, Grammatical Inference.
\end{IEEEkeywords}

\section{Introduction}
In~2019, \textit{World Tourism Organisation} UNWTO\footnote{International tourism Growth continues outpacing the global economy: edition 2020.} recorded $1.5$~billion international tourists, $4\%$ more than the previous year.
Tourism is one of the most important areas of the world economy. It is also considered to be one of the fastest-growing industries in the world~\cite{cooper2007contemporary}.
In this economic context, the understanding and knowledge of travel motivations and the anticipation of tourists' behaviors play an essential role in tourism marketing. It can lead to the recognition of demand, to make targeted advertising~\cite{Jingjing2018}, and help tourists to make decisions~\cite{march2005tourism}.

With the recent booming of digital tools and mobile internet technology, alternative sources of data to understand tourism behavior have emerged. Users of social networks, like $Tripadvisor$, $Booking$, $Facebook$, $Instagram$, tend to share openly and frequently photos, reviews, recommendations and videos of tourist places. Thus, when users share photos or reviews, geographical information is included. These geo-located data represent tourism and sociological views~\cite{chalfen1979,chareyron2009}.

Analyzing the behavior of tourists represents an important challenge to better monitor their movement and spreading in a given area. We can subsequently adapt supply to demand, recommend a stay to a tourist, provide relevant information for the tourism industry and management. In this paper, we focus our reviews on the prediction of tourist movements.

This paper addresses the problem of modeling and predicting the future movement of a tourist based on his present and past practices in a given area. Based on the geo-localized and temporal data information, we propose in this paper a model for predicting future tourist movement by analyzing the time sequences of places visited by a set of tourists. Our approach is to learn tourist's practices from a set of temporal sequences of places, through various methods to handle the difficulties due to big data.
The proposed method uses a machine learning method building a Hidden Markov model representing the whole data set. The model can produce predictions as a recommendation of future places to visit. The model can also be updated to adapt to new data.

Our model is built from the whole data without reducing its size and extracting a mathematical model. Thereby, we propose a new algorithm for automatic learning of grammatical inference to reduce its complexity in the context of big data. Moreover, this algorithm is designed to maintain all behavioral possibilities on the data set.
The principal contributions of our works are:
\begin{enumerate}
\item A method to establish sequences representing a unique stay of a tourist from a data set;
\item A new method of grammatical inference for processing very large data set;
\item A flexible and relevant decision-making tool to represent all tourists' movements in a data set. This decision-making tool is able to predict future visits as recommended places to visit.
\end{enumerate}

To validate the method, the results will be compared to the data statistics of the data set. Our method is not compared to deep learning methods of the literature.

In this paper, we will first describe in Section~\ref{Related work} the related work on network analysis. In section \ref{Data Analyse} presents data analysis methods, and in section \ref{Our approach} provides the various process of our method.
Our model is implemented and is the subject of a case study introduced in section \ref{Case study}. In section \ref{Conclusion}, we conclude our paper.

\section{Related work}
\label{Related work}

Many data analysis methods and algorithms are used in social networks to study and understand interactions between the entities, to predict the presence of connections, or to predict the following movement of an entity \cite{knoke2019social}.
Each method uses specific theories to produce an understanding of the network. 
To generate predictions of future movements in the networks, many studies typically use techniques and concepts from various fields of research like Machine Learning, Data Mining and Stochastic Models. These fields are not independent, studies combine methods and theories from these fields to analyze and to predict behavior. Related work is structured in these three axes by their principal contributions.

\paragraph{Machine learning methods}

Machine learning is a broad area of research that focuses on identifying patterns in empirical data. Machine learning algorithms establish a mathematical model based on sample data, known as training data, to make predictions or decisions. The sample data representing temporal sequences and are seen by machine learning algorithms as time series~\cite{timeSeries}.
In social network analysis, time series analysis is widely used for a variety of purposes. Many studies are working on social networks to predict relations between elements to be recommended by exploring the evolution of topological metrics~\cite{da2012time}.
Tourism research has taken a leap forward with the development of deep learning like neural networks~\cite{Askari2020,Zhan2018}. They are robust to noise in input data, in the mapping function and can even support learning and prediction in the presence of missing values. \textit{Wang} et al.~\cite{wang2004predicting} presents a model based on a neural network to estimate tourist arrivals in $Taiwan$ from $Hong~Kong$, $United~States$, and $Germany$. \textit{Zhang} et al.~\cite{zhang2019discovering} propose a deep learning analysis of photos to define tourists' behaviors in $Beijing$, $China$.

Machine learning and deep learning methods show their effectiveness over long time series containing thousands of points.
With short time series, composed of less than five elements which is the case in our study, classic machine learning methods fail to provide a relevant model. The primary drawback of machine learning, especially in deep learning is the difficulty to train the model. The model can oscillate between a local generalization and an extreme generalization.  
It is equally complex to determine an efficient deep learning structure because of significant variations in tourists' movements depending on the studied places, the source of the data, and the objectives of the study.

\paragraph{Data mining methods}

A second approach to generate a prediction from a data set is to use data mining methods. Data mining is a set of techniques that allow valuable information to be extracted from vast and loosely structured databases. To produce a prediction in a time-related data set, data mining methods need to decompose it into sequences.
Data mining methods provide information about how an element is represented in the data set or information about relations between each element.

Related to tourists' behavior extraction, various studies combine machine learning methods with data mining. \textit{Bao} et al.~\cite{bao2020bilstm} predict the next location to be visited by a tourist in $Wuhan$, $China$. This approach is based on a clustering area graph of user preferences. \textit{Majid} et al.~\cite{majid2015system} use weather and Points of Interest in Flickr to recommend the next location to visit in $Beijing$, $China$, thanks to pattern mining and similarity matrix. \textit{Memon} et al.~\cite{memon2015travel} presents a probabilistic approach for the recommendation based on various context rank and kernel estimation of Flickr geotagged data. The object is to determine the next position of the tourist in various cities of $China$. \textit{Wan} et al.~\cite{wan2018hybrid} employed a Bayes classifier to predict a user's latent interest in a location, based on weather, date, geotagged data of Flickr in the region of $Beijing$.

Data mining has two significant drawbacks for tourists' next visits recommendation. First, data mining methods use thresholds due to their complexity. Thus, only the most common behaviors are kept as results of data mining methods.
Second, most methods deal with the temporal order in a sequence but not with the relationship between elements according to their position in the sequence.

\paragraph{Stochastic model}
Another area of research to predict the next visits is to represent the data set of the sequences with a stochastic model. A stochastic model, or a Markov process, is a random process indexed by time, and with the property that the future is independent of the past, given the present. A finite discrete Markov chain is represented as a weighted oriented graph~\cite{howard1960dynamic}. An arc between two nodes represents the probability of reaching the final node as the next element in a sequence.

When it comes to recommendation or prediction, Markov process models show promises~\cite{HuyQuanVu2015,Saadi2016,Wang2016}. Among relevant studies, \textit{Ahmad} et Al.~\cite{Ahmad2019} propose a Markov model and a steady analysis to create a popularity index based on historical trajectories of mobile visitors inside $Jeju~Island$, $South~Korea$. \textit{Xia} et Al.~\cite{xia2011spatial} proposes a semi-Markovian process to model the probability that a tourist will visit and spend a certain amount of time in each location on $Philip~ Island$, $Australia$. In~\cite{RAMEZANI20121576,yeon2008travel}, the Markov process is used for arterial route travel time distribution and road congestion prediction. A variant of the Markov model was also used for the popularity of the routes~\cite{hazelton1996}.

Those methods resolve both of the drawbacks of data mining and machine learning mentioned above but are very sensitive to the distribution of elements in the sequences. Since machine learning, data mining, and stochastic models have pros and cons, we propose a hybrid method to learn tourist behaviors. This method is composed of a data analysis defining the concept of tourist and tourist stays. From these stays, we use a new machine learning method providing a prediction model. This model can be adapted to new stays and can predict behaviors that reflect the data set.

\section{Data Analysis}
\label{Data Analyse}

Our database is composed of users, reviews, and geo-located locations. A location is composed of coordinates $(lat, long)$ and a rating. To characterize the location, each of them has been aligned with administrative areas (GADM)\footnote{GADM:\url{https://gadm.org/index.html}. $386,735$ administrative areas (country, region, department, district, city, and town).}.
A user is identified by nationality, age and is described by a timeline. A user timeline represents a chronological set of reviews from its first reviews to its last reviews. This timeline allows computing intermediate properties: the time between two consecutive reviews, consecutive visited places.
A review represents a note given by a user on a location at a given discrete-time domain.

Now we must clarify the key points of our study, the definition of a tourist and a tourist's stay.

\subsection{Tourist}

According to the UNWTO, a tourist is defined as:
\textit{"The activities that people undertake during their travels and their stays in places outside their usual environment for a consecutive period not exceeding one year from recreational, business and other purposes".}

The term \textit{activities} should be understood here in a general sense of individual occupations.

\subsection{Tourist stay}

The main objective of our study is to model and predict tourist next visits over a given area by analyzing past and present individual tourist movements. The model is established on a set of \textit{sequence} created from the tourists' stays.
A tourist stay is a succession of days where a tourist publishes at least one comment per day. As soon as there is a day when the tourist does not post any comment, the stay is considered interrupted. However, a tourist may not post anything for a limited period of time during the same stay. Therefore, we determine the maximum number of possible days between two comments before considering it as a new stay.
According to the idea of \textit{Gossling}~\cite{Gossling2018}, we consider that reviews should be written within a maximum of $7$ days.

\begin{figure*}[!ht]
  \centering
    \includegraphics[width=0.8\textwidth]{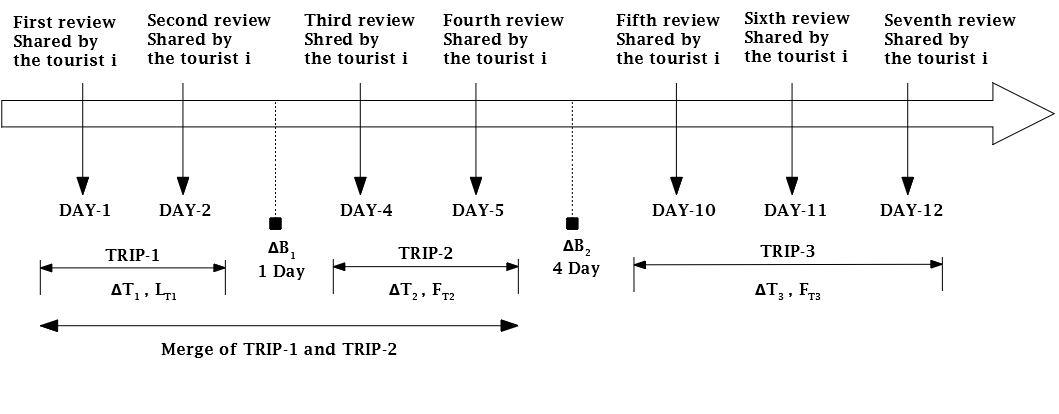}
    \caption{Example of merging two stays.}
    \label{merge trip}
\end{figure*}

The method consists of merging two stays if they satisfy the following conditions~\ifthenelse{\equal{\anonymized}{false}}{\cite{baccar2019tourist}}{[ANONYMIZED]}:
$$ \Delta{B} \leq \Delta{S_i} \; and \; \Delta{B} \leq \Delta{S_j} \; and \; L_{S_i} = F_{S_j}  $$
where $\Delta{B}$ is break duration, $\Delta{S_i}$ is the $i^{th}$ stay duration $\Delta{S_j}$ is the $j^{th}$ stay duration, $L_{s_i}$ is the last area visited during $i^{th}$ stay and $F_{S_j}$ is the first area visited during $j^{th}$ stay. 
The Figure~\ref{merge trip} presents an example of merge two stays  \textbf{Stay-1} and \textbf{Stay-2}.

Based on a user timeline, several stays are defined. Each one is represented by a list of visited places ordered temporally in a \textit{sequence}.
Our data set is a set of temporal sequences. Each sequence is composed of places, placed in the chronological order of appearance on the tourist's timeline, thus in the order of the stay.  

\section{Our approach}
\label{Our approach}

The purpose of this paper is to use temporal sequences to predict the next visits of tourists.

To make the prediction, we decompose our approach into three steps:
\begin{itemize}
\item[-] \textsc{Learning process}. This first step consists to understand tourist movements over a set of sequences and then learn past and present movements to predict a future one. The learning process, named \textit{Grammatical Inference}, consists of inducing rules, called \textit{Grammatical Rules}, from the sequence set. From these, a \textit{Hidden Markov Model (HMM)} is built. An HMM is a graph with nodes and arcs weighted by probabilities.
\item[-] \textsc{Prediction}. The second step is based on the Hidden Markov Model. Given a node of the HMM, all the future movements are generated browsing the HMM.
\item[-] \textsc{Update}. The third step is to update and adapt the weighted nodes and arcs of the Hidden Markov Model to a given set of sequences.
\end{itemize}

\subsection{Learning process}

The learning process is a set of algorithms to develop a mathematical model from raw data in aims to induct knowledge, called \textit{Grammatical Rules}.
To learn from the sequences' set and to induce grammatical rules, our learning process starts by modeling the sequences set in the \textit{Frequency Prefix Tree}. Based on the tree model, grammatical rules are inducted. To make predictions easier, we have structured the rules in an HMM.
In what follows, we will represent each element of the sequence by a unique integer called an item. In other words, an item refers to a place visited by a tourist.

\subsubsection{Frequency Prefix Tree (FPT)}
FPT also known as $Trie$, represents a directed and rooted hierarchical tree where the nodes and the arcs are weighted.  

The arcs between two nodes of FPT are multi-weighted by two attributes; 1) an  $item$, represented in parenthesis and 2) the $occurrence~frequency$ of the item in the sequences set.  

The nodes of FPT have one attribute that corresponds to the number of times a sequence ends in this node. By default, it equals zero.

\paragraph*{Construction of FPT} The process of constructing FPT is described in Algorithm~\ref{FPT}. Initially, the FPT is only composed of a root node with the attribute equal to zero. The construction of FPT consists of inserting each item of the sequences in the tree. The insertion process is an iterative process on each sequence as follows:

\begin{itemize}
    \item Starting at the root of the tree, the tree traversal is performed to insert each item of sequences
    \item The traversal compares the first item of the sequence to the arcs attribute item in the root.
    \begin{itemize}
         \item If no arc exists or no arc has the same item attribute, a new arc to a new node is created. The attribute item of the new arc will be the item of the sequence and a frequency equals to one.
    
         \item If an arc with the same attribute item is found, its frequency is incremented by one. Then, the traversal searches the next item of the sequence in the corresponding sub-tree. %
         \item When an item is the last of a sequence, then the attribute in the end node of the traversal is incremented by one.
    \end{itemize}

\end{itemize}

\begin{algorithm}[!ht]
\SetAlgoLined
\KwIn{sequences set $A$}
\KwOut{Return FPT from sequences set $A$}
 Create root node of the FPT\;
 \ForEach{Sequence $a$ in $A$}{%
 node 	$\leftarrow$ root \;
 \ForEach{Item $l$ in sequence $a$}{%
 \uIf{node has arc with item $l$}{
 arc($l$).frequency $\leftarrow$ +1\;
 }
 \Else{
  node.addArc($l$)  {\footnotesize // Create an arc and its end node}
 arc($l$).frequency $\leftarrow$ 1\;
 }
  node $\leftarrow$ getArc($l$) {\footnotesize // Go the the end node of an arc} \\
  \If{$a$ is end of sequence}{node.sequence\_ends $\leftarrow$ +1}
  }
 }
 \Return root
 \caption{Construction of a FPT.}
 \label{FPT}
\end{algorithm}

The generated FPT cannot be used directly due to its large number of nodes. The number of nodes is at most the number of unique items power the length of the largest sequence in the sequences set. To overcome this problem, one must reduce its size using \textit{grammatical inference (GI)}.

\subsubsection{Grammatical Inference (GI)}

The grammatical inference is a machine learning method that consists of reducing an FPT into a \textit{Stochastic Automaton} by merging nodes.
Two nodes can be merged if they are compatible \textit{i.e.,} they valid the compatibility test, see definition \ref{def:def1}.

\begin{definition}[Compatibility test]
\label{def:def1}
Two nodes are compatible if all their arcs to their children until reaching the leaves have the same item attribute with the same \textit{Relative Frequency}, see definition \ref{def:def2}, following the \textit{Hoeffding bounds}.
Parameters of Hoeffding bounds follow Mao et al. recommendation~\cite{mao2016learning} and described by \textit{Thollard} et al.~\cite{thollard2000probabilistic}.

\end{definition}

\begin{definition}[Relative Frequency]
\label{def:def2}
\begin{itemize}
    \item [] \textbf{Relative frequency of an arc} is equal to its frequency attribute divided by the sum of frequencies attributes of the outgoing arcs of its start node; \textit{i.e.,} the outgoing arcs of its parent.
 \item [] \textbf{The relative frequency of a node} is equal to its attribute divided by the sum of all ingoing arc frequencies to that node. For the root, it is divided by the sum of the outgoing arc frequencies.

\end{itemize}

\end{definition}

In the literature, there are three mains GI algorithms, \textit{MDI}~\cite{carrasco1994learning}, \textit{Alergia}~\cite{thollard2000probabilistic} and \textit{DEES}~\cite{habrard2006using}. 
In our approach, we use the \textit{Alergia algorithm} because of its error function. The error function computes the rate of information loss during the GI merging process.

\paragraph*{Alergia algorithm} is a GI method, which takes as a parameter the FPT and iteratively attempts to merge all node pairs if they are compatible. 
With Alergia algorithm, the compatibility between two nodes depends on the compatibility between all their children until the leaves. 
To check if two nodes are compatible, the number of tests to compute with a given FPT having $n$ different items and a height of $h$ is at most $n^h$. Performing the compatibility test between each pair of nodes in the FPT generates an explosion of combinations.

To counter this problem, we purpose a \textit{Relaxed Alergia algorithm}. 
With the Relaxed Alergia algorithm, two nodes are compatible if and only if the arcs to their children have the same relative frequency following the \textit{Hoeffding bounds}.
Despite the difference of the compatibility test between Alergia algorithm and Relaxed Alergia algorithm, they return the same result if in FTP all children of two compatible nodes have a similar relative frequency on the same item.

\paragraph*{Relaxed Alergia algorithm} tries to recursively merge all the pairs of nodes, from the root to the leaves. 
To differentiate nodes that cannot merge from those that will be tested, we define two types of nodes, RED nodes and BLUE nodes.
In the beginning, only the root node of the FPT is RED, and BLUE contains the direct children of the root. The BLUE nodes are visited successively. Once visited, a BLUE node tries to merge with the RED node. If the visited node does not merge, it becomes a RED node and all its children become BLUE. This process is repeated until all FPT nodes are RED.

When a BLUE node is compatible with a RED node, the \textit{merging process} is launched. The merging process is composed of two operations \textit{Merge} (definition \ref{def:def3}) and \textit{Fold} (definition \ref{def:def4}).

\begin{definition}[Merge operation]
\label{def:def3}
To merge a BLUE node into a RED node, all the ingoing arcs of BLUE node are redirected to the RED node. If the RED node already has an ingoing arc with the same item attribute as a redirected arc, then only the frequencies are added. Next, the BLUE node attribute is added to the RED node attribute.
\end{definition}

\begin{definition}[Fold operation]
\label{def:def4}
When a BLUE node merges into a RED node, all children of BLUE node are recursively merged into children of the RED node with the same item attribute arc. If the arc with an item attribute doesn't exist in the RED node, a new child is inserted.
\end{definition}

\begin{figure}[!ht]
\centering
\subfloat[Initial graph.\label{fig:Initial}]{%
  \includegraphics[width=0.30\linewidth]{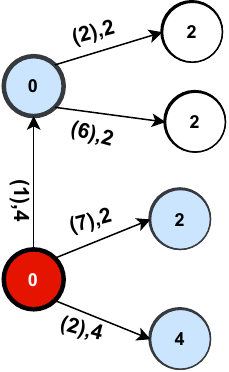}%
}\hfil
\subfloat[Merge operation.\label{fig:Mergedone}]{%
  \includegraphics[width=0.30\linewidth]{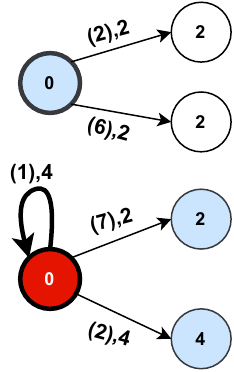}%
}
\hfil
\subfloat[Fold operation.\label{fig:folding}]{%
  \includegraphics[width=0.30\linewidth]{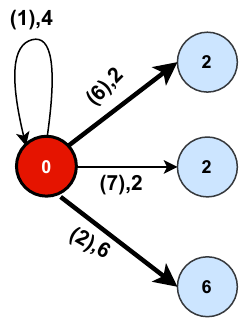}%
}
\caption{Merge and fold operations of the Relaxed Alergia algorithm.}
\label{fig:merge}
\end{figure}

The Figure~\ref{fig:merge} presents an example of the merge and fold process.
We consider that the RED node and the top BLUE node are compatible, see Figure \ref{fig:Initial}. The RED node merges with the BLUE node in bold. The ingoing arcs of BLUE with attributes $(1),4$ is redirected to the RED node then make a loop, see Figure \ref{fig:Mergedone}. The fold operation is on the two children of BLUE, see Figure \ref{fig:folding}. The item $(2)$ in the arc with attributes $(2),2$ already exists in RED as $(2),4$, thus the arc becomes $(2),6=2+4$. The node with attribute at $4$ adds the attribute $2$ thus becomes $6$. The item $(6)$ in the arc $(6),2$ doesn't exist in RED, thus a new child with the same attributes is created.

The output of the Relaxed Alergia algorithm is a graph with frequency and item called \textit{frequency automaton}.
To make a prediction, in the literature, there are no existing algorithms or methods that allow us to analyze a frequency automaton. Thus, we will transform the frequency automaton into an \textit{Hidden Markov Models}. 

\subsubsection{Hidden Markov Models}

To transform a frequency automaton into a Hidden Markov Models (HMM), one has to start to change the frequency automaton into a \textit{stochastic automaton}. This first step consists to convert the frequencies to relative frequency. 

\begin{figure}[!ht]
\centering
\subfloat[Stochastic~automaton.\label{stoch}]{%
  \includegraphics[width=0.30\linewidth]{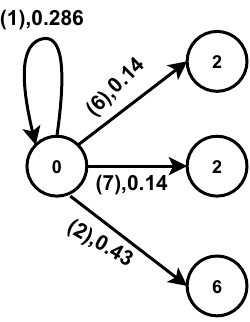}%
}\hfil
\subfloat[Corresponding HMM.\label{HMM}]{%
  \includegraphics[width=0.70\linewidth]{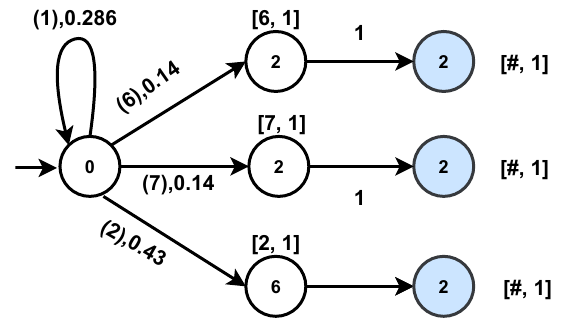}%
}
\caption{Changing a stochastic automaton into an HMM.}
\label{fig:image2}
\end{figure}

The Figure~\ref{stoch} provides the stochastic automaton of the frequency automaton of Figure~\ref{fig:folding}. The root node has four outgoing arcs: $(1),4$; $(6),2$; $(7),2$; $(2),6$. The sum of frequencies is equal to $4+2+2+6=14$, thus the arcs in relative frequencies are: $(1),0.286$; $(6),0.14$; $(7),0.14$; $(2),0.43$.

The process to transform a stochastic automaton into an HMM is described and proved by \textit{Dupont} et al.~\cite{dupont2005links} and \textit{Harbrard} et al.~\cite{habrard2006using}. The steps of the process are not discussed in our paper. Although we provide an explanation of the structure of an HMM.

An HMM is a graph where a node has a set of couples $[item, probability]$, refer to the $probability$ to generate the $item$ on this node. Note the item $\#$ refers to the end of a sequence. The root node of an HMM is characterized by an ingoing arc without a start node. Arcs possess a $probability$ to go from a node to another node. 

To be able to predict, we introduce two operations on an HMM: \textit{jump} and \textit{observation}.
The jump operation consists of going from a node to another one through an arc. After a jump, one can generate an $item$ given its $probability$. The jump and item generation form an observation. The probability to observe this item is equal to the product of the two probabilities, \textit{i.e.,} the jump and the item's generation.
The Figure \ref{HMM} illustrates the transformation of the stochastic automaton in Figure \ref{stoch} to an HMM. 

The probabilities of outgoing arcs in the root stochastic automaton are : $(1),0.286$; $(6),0.14$; $(7),0.14$; $(2),0.43$. Let's observe item $(1)$ in the HMM. This is only possible by jumping through the loop and generating an item on the root. 
Thus, the probability to observe $(1)$ from the root is equal to $0.286*1=0.286$.

\subsection{Prediction}

In an HMM, the prediction process is done from a given sequence of items. Therefore, it is necessary to find where to start the prediction thanks to the \textit{Viterbi algorithm}~\cite{kriouile1990some}. Given an input sequence of items, this algorithm computes the most probable sequence of observations.
Hence, the prediction process occurs at the node of the last observation of this sequence.
A set of all predicted observations is generated from this node. This means all jumps from the start node to another node and all generated items at these nodes are computed.

An observation is composed of couples \textit{(suffix, probability)} where $suffix$ refers to the generated item with its $probability$ to be observed.
If we want to continue to predict new places to visit from a given $suffix$, all new observations are computed. In the set of new observations, the new $suffix$es are concatenated to the old $suffix$. Since many observations can build the same new $suffix$es, their probabilities are added. The prediction process is shown in Algorithm~\ref{prediction}.

In our context, we want to predict the next visited places of a tourist given a chronological sequence of the past and present visited places. Following the prediction process, we can recommend visiting one or many $suffix$es following the decreasing value of their probabilities.

\begin{algorithm}[!ht]
\SetKwInOut{Input}{Input}
\SetKwInOut{Output}{Output}
    \Input{An HMM with $O_{i,l}$ the probability to observe element $l$ on node $i$\\
    $P_{i,j}$ the probability to jump from node $i$ to\\
    node $j$ at a discrete time\\
    Length of suffix $L$\\}    
    \Output{A set $S_O$ of observations}
    \SetKwProg{Fn}{Function}{ is}{end}
    $S_O$ $\leftarrow$ empty\;
    root $\leftarrow$ End point of \textit{Viterbi(s)}\;
    \textit{Suffixes(root, 1, $S_O$, empty, $L$)}\;
    \Fn{Suffixes(root, probability, set, suffix, $L$)}{}
    {
    \uIf{$L==0$}{
    \Return\;
  }
  \Else{
    \ForEach{Child $c$ in root.children}
    {
    \ForEach{Observation $l$ in $c$}
    {
    suffix $\leftarrow$ concat(suffix,'l')\;
    probability $\leftarrow$ probability*$P_{root,c}$*$O_{c,l}$\;
    $S_O$.add(suffix,probability)\;
    \Return \textit{Suffixes($c$, probability, $S_O$, suffix, $L-1$)}\;
    }
    }
    }
    }
    Sum of probabilities of same suffixes in $S_O$\;
    \caption{Prediction Process.}
    \label{prediction}
    \end{algorithm}

\subsection{Update} 

The update process adapts the probabilities to jump and the probabilities to generate items for a sequence set. To update the HMM, we use the \textit{Baum-Welch algorithm}~\cite{baggenstoss2001modified,kriouile1990some}. In our method, we implement this algorithm without any modification, so it isn't discussed in this paper.

\section{Case study}
\label{Case study}
The proposed method is implemented with data extracted from $Tripadvisor$, concerning posted reviews in $Paris$, the capital of France, from 2013 to 2019. 
For this case study, we consider only the 6 most visited places of Paris. In what follow, the places are renamed as: $(0)$ represents $Arc~de~ Triomphe$, $(1)$ $Notre~Dame~Cathedral$, $(2)$  $Eiffel~Tower$, $(3)$ $Luxembourg's~Garden$, $(4)$ $Quai~de~Seine$ and $(5)$ $Louvre's~Museum$.

After data processing, we extract from $1'063'447$ reviews $11'471$ sequences of at least $2$ items. Indeed, sequences of only $1$ item are removed because they do not show the movement of a tourist. Based on the sequences set, the generated FPT has $599$ nodes and $556$ of them have a non-null attribute. 
The complete source code and detail of results and figures are provided in a GitHub folder\footnote{https://github.com/TheoDemessance/Hidden-Markov-Model-for-Tourism}.

\subsubsection{Validation of the Relaxed Alergia algorithm}
First of all, we check if in FTP all arcs with the same item attribute have a similar relative frequency.
The results show all the children with the same item from a given sequence have relative frequencies with an absolute variation of at most $8\%$. The Relaxed Alergia algorithm can be used on the sequences set. Therefore, it exists three anomalies: item $(5)$ at the first position and second position, and item $(3)$ at the first position.
Due to the anomalies, the HMM model will provide some probabilities on the sequences containing those items with a more significant error than the other sequences comparing to the sequences set.
Applying the Relaxed Alergia algorithm, the generated HMM has $37$ nodes with $18$ end nodes.
To compensate the three anomalies, the HMM will be updated with the sequences set many times before beginning the predictions. To prove the efficiency of the update process, we will compare the HMM before and after updating in what follows.

\subsubsection{HMM's validation}
To validate our HMM, we verify if it's a reflection of the sequences set. Let's compare the probabilities of a sequence in the sequences set to the probability of the same sequence observed in the HMM. %
For a given sequence in the sequences set, we use the \textit{Baum-Welch algorithm}~\cite{baggenstoss2001modified,kriouile1990some} to compute its probability to be observed in the HMM. 
To compare the probability $R_s$ of a sequence $s$ in the sequences set to its probability $P_s$ to be observed in the HMM, we use the \textit{Absolute Percent Error} (APE). APE is defined as follows:
$\mbox{APE}= \left| \frac{R_{s}-P_{s}}{R_{s}}\right|$.

\begin{figure*}[!ht]
  \centering
    \includegraphics[width=.9\textwidth]{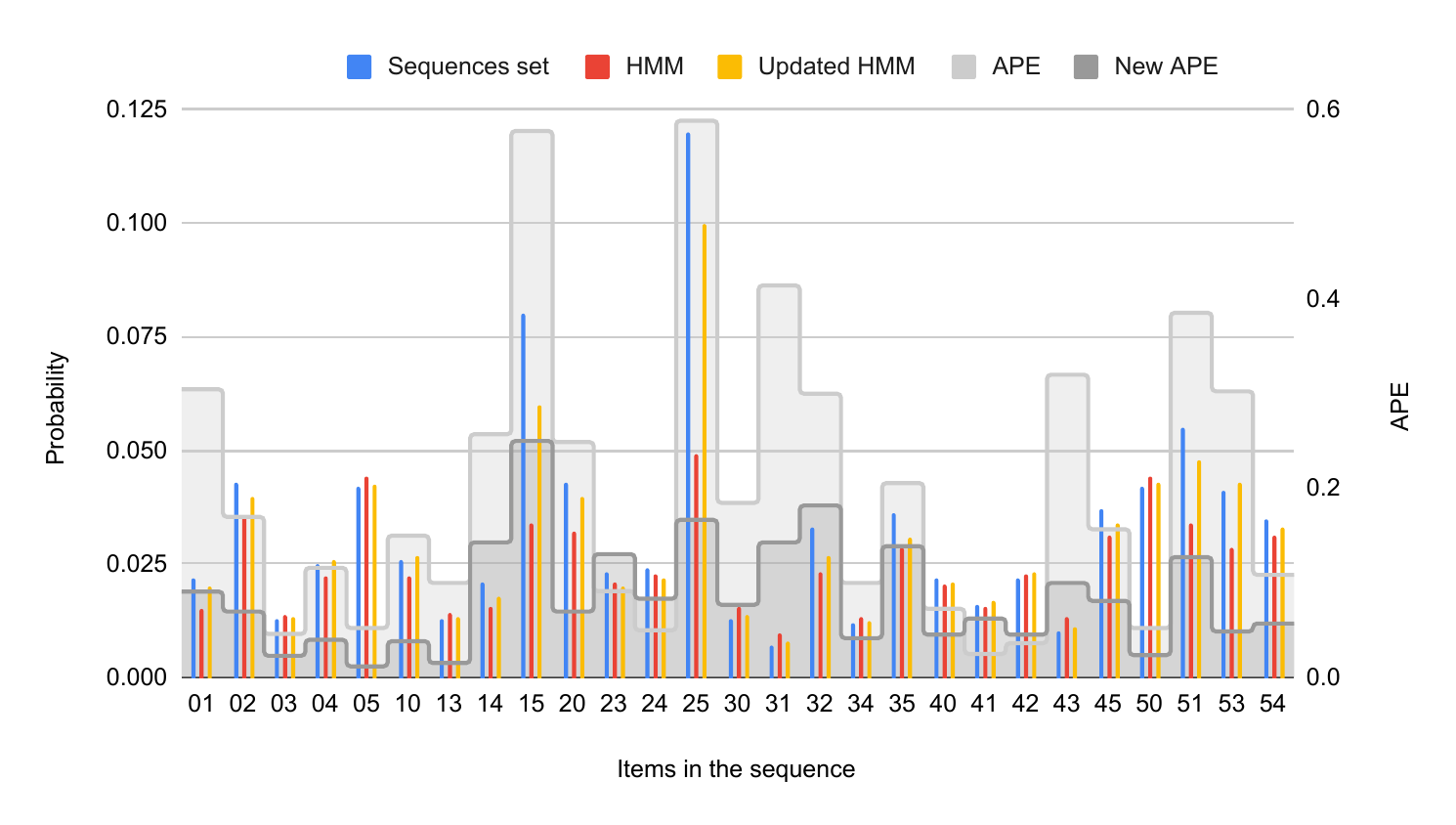}
    \caption{HMM's validation.} 
    \label{datacompfig2}
\end{figure*}

The Figure~\ref{datacompfig2} shows the APE between the probabilities from the sequences set and the observed probabilities in the HMM. In the x-axis, each digit represents an item. For example $05$ means the sequence $\{0,5\}$, \textit{i.e.,} \textit{{Arc de Triomphe, Louvre's Museum}}.
The APE values are between $3\%$ to $66\%$ with its mean (MAPE) equals to $20.8\%$.
In consequence, various observed sequences are not well represented in the HMM. The item $(5)$ in the first and second position and the item $(3)$ in the first position produce the largest APE values as expected. 

To provide a more accurate observation of the sequence containing those items, we update the HMM with the sequences set until the MAPE is strictly inferior to $10\%$. 
The figure \ref{datacompfig2} also presents the probabilities of observations of the update HMM with the new APE on each sequence. The MAPE is equal to $8.9\%$, APE is between $1.2\%$ to $25\%$. 
The APE significantly decreases for all sequences and they are more restricted around the MAPE. These results demonstrate how the HMM adapts itself to better reflect the sequences set. 

\subsubsection{HMM's prediction}
\begin{figure*}[!ht]
  \centering
    \includegraphics[width=.7\textwidth]{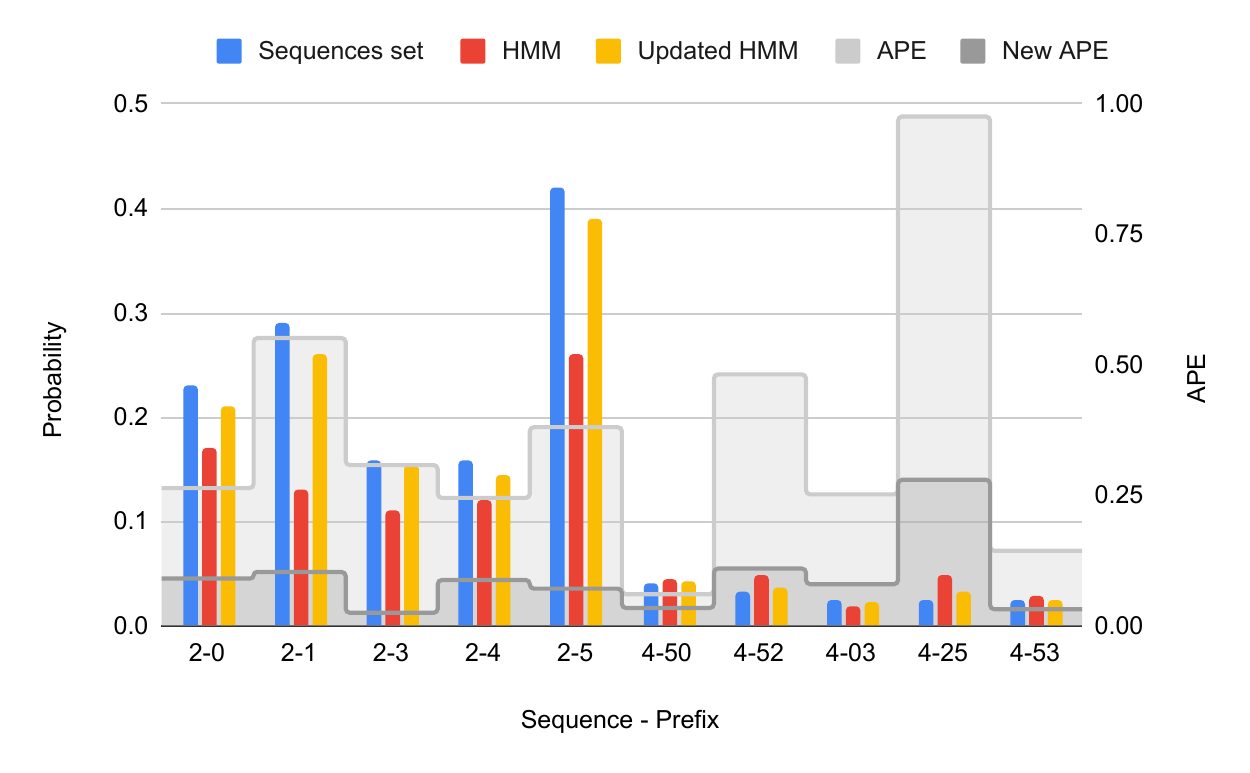}
    \caption{HMM's predictions.} 
    \label{datapred}
\end{figure*}

Presently, we can predict in the HMM the future behavior of a tourist based on a sequence of places visited.
The figure \ref{datapred} presents the probability to observe suffix in the HMM, in the updated HMM, and their respective APE.
The sequence is composed of two sub-sequences $A$-$B$ where $A$ is the sequence of places visited and $B$ is the predicted suffix.
The MAPE between the sequences set and the HMM is equal to $36\%$, with APE values from $6\%$ to $98\%$. In the updated HMM, the MAPE is equal to $9.1\%$, with APE value from $2.5\%$ to $28\%$.
The Relaxed Alergia algorithm has built an HMM that adapts to all sequences. The update process improves modeling and forecasting results.

The proposed model provides relevant and close to reality predictions of tourist behavior. Models can be constructed from a set of restricted sequences based on nationality, age, gender, etc. to constitute a model that fits a set of tourists. To make recommendations for a tourist, a set of suffixes is built from its visited places. One can recommend the suffix with the largest probability or a suffix containing the next visits with the largest probability.

\section{Conclusion}
\label{Conclusion}

The main idea of this study is to determine a decision tool representing the movements of a set of tourists. Our proposed method keeps all the possible movements. When a latter is over-represented or under-represented, our method adapts the decision tool to better fit the reality.

Movements may vary according to the country of origin, age, gender, social class of a set of tourists. Since the methods presented can predict the next visits of a specific group of tourists, we will compare the decision tool of various sets of tourists in future work. We aim to provide different decision tools according to the profile of tourists. Moreover, when a tourist needs a recommendation, his best profile is determined to best meet his expectations. 
In addition, to validate our approach, we will apply deep learning methods to compare results.

\bibliographystyle{IEEEtran}
\bibliography{biblio}

\end{document}